\documentclass[conference]{IEEEtran}
\IEEEoverridecommandlockouts
\usepackage{cite}

\usepackage{algorithmic}

\usepackage{textcomp}
\usepackage{xcolor}
\usepackage[pdftex]{graphicx}
\usepackage[cmex10]{amsmath}
\usepackage{pgf-pie,etoolbox}
\usetikzlibrary{shapes}
\usepackage{tikz}

\makeatletter
\patchcmd\pgfpie@slice
  {node {\scalefont{#3}\beforenumber#3\afternumber}}
  {node[/every only number node/.try] {\scalefont{#3}\beforenumber#3\afternumber}}{}{}
\usepackage{amsmath}
\usepackage{placeins}
\usepackage{caption}
\usepackage{mdwtab}
\usepackage{graphicx}

\usepackage{eqparbox}
\usepackage{titlesec}
\renewcommand{\arraystretch}{1.6}

\usepackage{url}
\usepackage{float}
\def\BibTeX{{\rm B\kern-.05em{\sc i\kern-.025em b}\kern-.08em
    T\kern-.1667em\lower.7ex\hbox{E}\kern-.125emX}}
\begin{document}

\title{End-to-End Optimized Pipeline for Prediction of Protein Folding Kinetics\\
{}
}

\author{\IEEEauthorblockN{1\textsuperscript{st} Vijay Arvind.R}
\IEEEauthorblockA{\textit{Department of Computing Technologies,} \\
\textit{Faculty of Engineering and Technology,}\\
\textit{SRM Institute of Science and Technology,}\\
Tamil Nadu-603 203, India \\
va0149@srmist.edu.in}
\and
\IEEEauthorblockN{1\textsuperscript{st} Haribharathi.S}
\IEEEauthorblockA{\textit{Department of Computing Technologies,} \\
\textit{Faculty of Engineering and Technology,}\\
\textit{SRM Institute of Science and Technology,}\\
Tamil Nadu-603 203, India \\
hs7886@srmist.edu.in}
\and
\IEEEauthorblockN{2\textsuperscript{nd} Brindha.R}
\IEEEauthorblockA{\textit{Department of Computing Technologies,} \\
\textit{Faculty of Engineering and Technology,}\\
\textit{SRM Institute of Science and Technology,}\\
Tamil Nadu-603 203, India \\
brindhar1@srmist.edu.in}
}

\maketitle

\begin{abstract}
Protein folding is the intricate process by which a linear sequence of amino acids self-assembles into a unique three-dimensional structure. Protein folding kinetics is the study of pathways and time-dependent mechanisms a protein undergoes when it folds. Understanding protein kinetics is essential as a protein needs to fold correctly for it to perform its biological functions optimally, and a misfolded protein can sometimes be contorted into shapes that are not ideal for a cellular environment giving rise to many degenerative, neuro-degenerative disorders and amyloid diseases. Monitoring at-risk individuals and detecting protein discrepancies in a protein's folding kinetics at the early stages could majorly result in public health benefits, as preventive measures can be taken. This research proposes an efficient pipeline for predicting protein folding kinetics with high accuracy and low memory footprint. The deployed machine learning (ML) model outperformed the state-of-the-art ML models by 4.8\% in terms of accuracy while consuming 327x lesser memory and being 7.3\% faster. 
\end{abstract}


%
\IEEEpeerreviewmaketitle


\section{Introduction}

Even though many debilitating diseases such as Alzheimer's disease (AD) and Parkinson's disease (PD) have catastrophic consequences for our human bodies, there is still no significant cure or development in early detection methods for these diseases. This significantly underscores the urgency of understanding the underlying principles behind these diseases. The people who suffer from AD not only have physical symptoms but are also affected emotionally. The inability to think, communicate and perform various tasks daily is often experienced by people suffering from AD \cite{grabher2018effects}. On the other hand, PD impacts one's movements and causes anxiety and depression \cite{contreras2003effects}. Prior research \cite{stefani2003protein} has nailed down that several disruptions in the protein folding process can lead to misfolded proteins, which form insoluble long linear or fibrillar aggregates in several body parts. This process is identified to be the primary cause of AD and PD \cite{m2014protein}.

Protein folding kinetics (PFK) becomes vital for studying protein dynamics and behavior. The PFK's magnitude reflects a protein's propensity to undergo these transitions between folded and unfolded states or the association and dissociation of protein complexes \cite{kiefhaber1995protein}. Various factors influence PFK, such as PH, temperature, electric and magnetic fields, etc. PFK can be determined using spectroscopy, Nuclear Magnetic Resonance (NMR), and fluorescence. However, these techniques often require specialized equipment and expertise. NMR experiments, for example, demand access to high-field spectrometers and advanced pulse sequences, which are not easily affordable.
Furthermore, spectroscopic methods, including UV-Vis and infrared spectroscopy, have limitations when capturing fast kinetic events due to slower data acquisition rates. Overcoming these limitations requires precise optimization of various parameters, development of specialized protocols, and integration of complementary techniques. Several advancements in instrumentation, data acquisition, and analysis methods are also intended to address these limitations and improve the sensitivity and precision of protein kinetic studies utilizing spectroscopy, NMR, and fluorescence techniques.

PFK predictions have been done prior to this research with machine learning and deep learning models. However, these approaches have encountered various difficulties due to the protein structure's intricate and irregular nature. These techniques have shown great potential in predicting, but their effectiveness is often inhibited by their inability to capture the fundamental nature of the protein structure between various parameters, thus being inaccurate \cite{corrales2015machine}.

\section{Related Work}
In the field of computational biology, numerous methods have been developed to predict protein folding kinetics (PFK). M. Michael Gromiha et al. \cite{10.1093/nar/gkl043} developed a machine learning model to predict the folding rates of proteins from their amino acid sequence. Their proposed method achieved an overall correlation of 98\% with an inference time of 8.44 seconds per protein for 77 two and three-state proteins from the Protein Data Bank (PDB). Chen-Chen Li et al. \cite{10.1093/bib/bbz133} proposed a fold-specific feature extraction method combined with a convolutional neural network. Their implementation produced an accuracy of 77\% overall. Balachandran Manavalan et al. \cite{MANAVALAN2022105911} constructed an ensemble of regression and classifier models to predict folding kinetics and fold type. Their constructed framework produced an accuracy of 84.3\%. David De Sancho et al. \cite{de2011integrated} proposed an algorithmic approach called PREFUR to predict PFK from the size and structural class of the protein. Although their method is unique, the experimental results only produced an accuracy of 70\%. Jianxiu Guo et al. \cite{guo2011predicting} developed an artificial neural network trained on 90 proteins. The presented implementation showed a correlation of 80\% and a standard error rate of 2.65. Although considerable research exists in PFK prediction, the previously proposed and implemented approaches are out-modded, highly inaccurate, and slow. \\
Saraswathy Nithiyanandham et al. \cite{NITHIYANANDAM2023106436} proposed a framework built on standard shallow machine learning models trained on PFDB to predict PFK. They explore several machine learning algorithms throughout their work and focus heavily on establishing relationships between the folding kinetics and other structural parameters. While their approach is commendable, they fail to implement an accurate predictive regressor by only achieving an RMSE of 3.030. 

This research proposes a flexible, optimized inference pipeline for protein folding kinetics prediction. The pipeline uses a lightweight ML regressor (bonsai) as the backbone trained on the most optimal feature subset derived from the base dataset, which consumes less than 0.7KB and has an incredibly low inference time (ms). In addition to optimizing the model for achieving the least inference time for embedded devices, feedback optimization techniques were incorporated to improve the performance iteratively while accommodating device limitations. The pipeline archives profound accuracy compared to other state-of-the-art (SoA) systems.

\section{Architecture}
\begin{figure}
  \centering
  \includegraphics[width=\columnwidth, height=5cm]{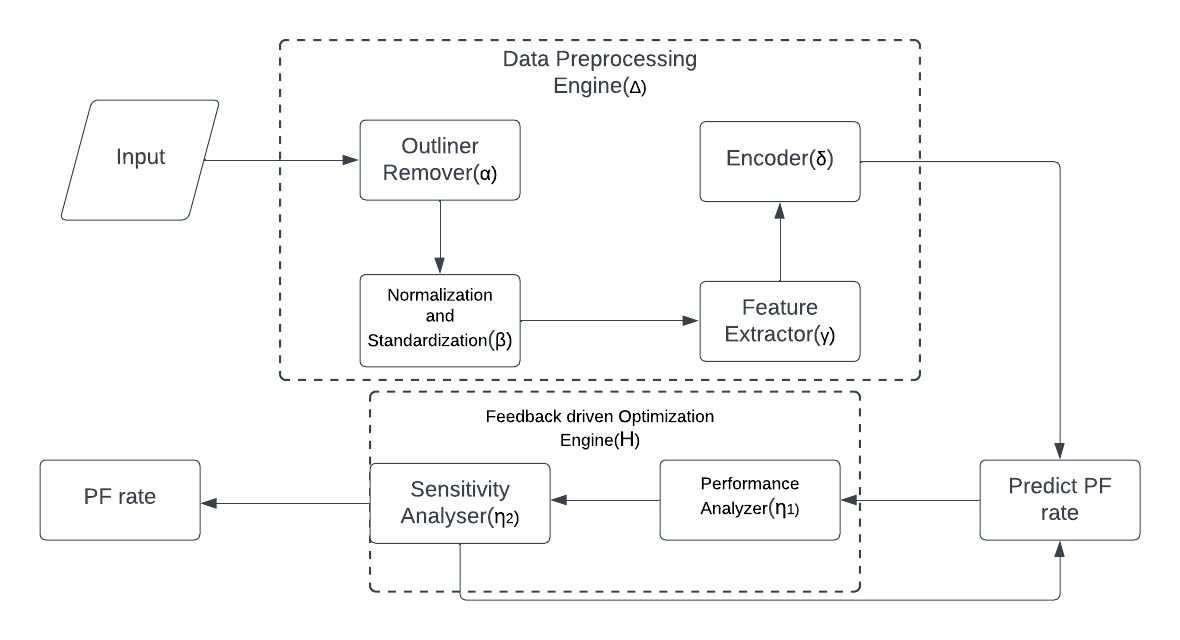}
  \caption{Pipeline Architecture}
  \label{pipe}
\end{figure}

The proposed pipeline is an end-to-end framework that takes in protein-specific parameters such as type of fold, amino acid sequence, torsion angles, etc, as its input and produces PFK predictions. The data is first fed into the pre-processing module to scale, clean, and normalize the data. The processed data is then pipelined to a tree-based optimized regression model. The need for a tree-based algorithm was ascertained due to its ability to form complex relationships among the parameters. The architecture of the end-to-end workflow is depicted in Fig~\ref{pipe}. An in-depth description of the iterative stages of the pipeline is outlined in the upcoming sections.
\subsection{Data \& Preprocessing}
Protein folding database (PFDB) \cite{manavalan2019pfdb} was considered for protein folding kinetics (PFK) prediction. PFDB consists of experimental observations for 2S (two adjacent strands of a beta-sheet) and N2S (N adjacent strands of a beta-sheet) proteins. Structural parameters such as amino acid sequence, type of chaining, torsion angles, Ramachandran outliers, etc., were manually derived from RCSB PDB \cite{10.1093/nar/28.1.235} for each protein. These structural parameters were considered because they significantly influence the type and nature of folding \cite{miller2002experimental}. Other individually reported observations of proteins were also collected and used for testing the regressor to test its real-time capabilities.

The data is first passed to the data pre-processing module, which consists of an Outlier Remover ($\alpha$), Temperature Standardization ($\beta$), Feature Extractor ($\gamma$), and Encoder ($\delta$). As the data consists of various features, $\alpha$ ensures the removal of outliers using the percentile-based outlier detection method, specifically the interquartile range (IQR) approach. This method establishes lower and upper thresholds based on the first quartile ($Q1$) and third quartile ($Q3$) values, respectively. The lower threshold is calculated using the formula $Q1 - 1.5 \times IQR$, while the upper threshold is determined as $Q3 + 1.5 \times IQR$. Data points falling above the upper threshold or below the lower threshold are classified as outliers, representing values that significantly deviate from the central distribution of the data.

The Temperature Standardization ($\beta$) step is applied for the temperature feature. $\beta$ standardizes the temperature value to an optimal scale. Temperature standardization is critical when dealing with protein folding data collected at various temperatures. $\beta$ collects the rate constants of folding and unfolding ($ln(k_f)$ and $ln(k_u)$, respectively) at a reference temperature of 25°C using the Eyring-Kramers equation \cite{bilsel2000barriers} which facilitates precise assessments of folding kinetics and enabling more robust conclusions in protein folding research. This correction adjusts the dataset to a standardized temperature, facilitating more accurate and reliable comparisons.

For the numerical features (\textit{Lpbd}, \textit{L}, \textit{pH}, and \textit{temperature}), $\beta$ applies Z-score normalization as the feature extraction technique. This process subtracts the mean and divides it by the standard deviation across each feature. Z-score normalization ensures that all numerical features are brought to a standardized distribution by eliminating differences in magnitude between them. Additionally, this enables more efficient and unbiased analysis.
\begin{table}[h]
\centering

\Large
\caption{Feature Set}
\scalebox{0.6}{
\begin{tabular}{|c|c|}

\hline
 \textbf{Feature} & \textbf{Description}  \\
\hline
Psn & Short name of the protein \\
\hline
Class & Class of the protein $(\alpha / \beta)$ \\
\hline
Fold & Type of fold classified by SCOP  \\
\hline
Lpdb & Number of continuous folded residues  \\
\hline
L & Total number of residues in the protein  \\
\hline
pH & pH level the protein was recorded \\
\hline
Temp & Tempterature the protein is recorded\\
\hline
$F_{type}$ & Fold type (N2s / 2S) \\
\hline
ln(ku) & Unfolding rate of the protein \\
\hline
$\beta$ T & Tanford $\beta$ value \\
\hline
\end{tabular}
}

\end{table}
\\
For the categorical features, $\gamma$ ensures that all categorical data points are encoded numerically. As mentioned earlier, the M-estimate encoder \cite{mougan2021quantile} is used for this purpose. The encoder calculates the probabilities of each column based on their frequency in the dataset. With a $m = 20$ value, the encoder effectively down-weights outliers and reduces their impact on the estimation process. 
\begin{equation}
    \Delta = [\alpha_,\hspace{+.2cm}\beta_,\hspace{+.2cm}\gamma_,\hspace{+.2cm}\delta]
\end{equation}

\subsection{Optimized Regression Model}
The machine learning algorithm utilized in this study is Bonsai \cite{kumar2017resource}, which employs a shallow tree-based approach to process data continuously and incrementally, tailored explicitly for constrained paucity systems. Data optimization and memory utilization is performed by learning the input data at a lower dimension. Bonsai's streaming implementation allows it to support devices with limited RAM, even those incapable of storing a single vector.  

These trees incorporate enhanced nodes, enabling internal and leaf nodes to make non-linear predictions. Bonsai achieves a remarkable ability to capture intricate non-linear decision boundaries by combining the predictions made by individual nodes along the path followed by a data point. In addition to enabling parameter sharing along paths, path-based prediction reduces the size of the overall model, thus increasing its efficiency. It optimizes memory allocation and maximizes prediction accuracy by learning all nodes jointly rather than node by node in a greedy manner. Bonsai offers an effective solution for machine learning tasks through its combination of sparse projection, streaming learning, enhanced nodes, and joint learning. This algorithm was specifically chosen for its low memory footprint, which would further help this pipeline to be integrated onto other fold-specific frameworks.
\subsection{Post Processing \& Aggregate Feedback}
Feedback-driven Optimization Engine (H) ensured iterative improvement of the pipeline over time. The Performance Analyzer $\eta_1$ assessed the quality of predictions, which consists of various evaluation metrics as discussed in Section~\ref{metric}. Sensitivity Analyzer $\eta_2$ facilitated a thorough observation of the impact of variations in input parameters and model architectures on the predictions, leveraging the results obtained from $\eta_1$, which compared the predictions against known ground truth values. Additionally, the testing data was split into different batches, and the accuracy was measured for each batch, aiding in the identification of trends and patterns in performance. With the help of $\eta_2$, various hyper-parameters were optimized to maximize the model’s performance. H is essentially needed for bonsai to boost its performance as the model primarily prioritizes minimizing the computational resources. Moreover, the iterative optimization process driven by H enabled continuous refinement and adaptation of the pipeline, leading to enhanced accuracy and reliability.

\begin{equation}
    \mathrm{H} = [\eta_1,\hspace{+.2cm}\eta_2]
\end{equation}

\begin{figure}
\subsection{Feature Importance}
  \centering
  \includegraphics[width=\columnwidth, height=5cm]{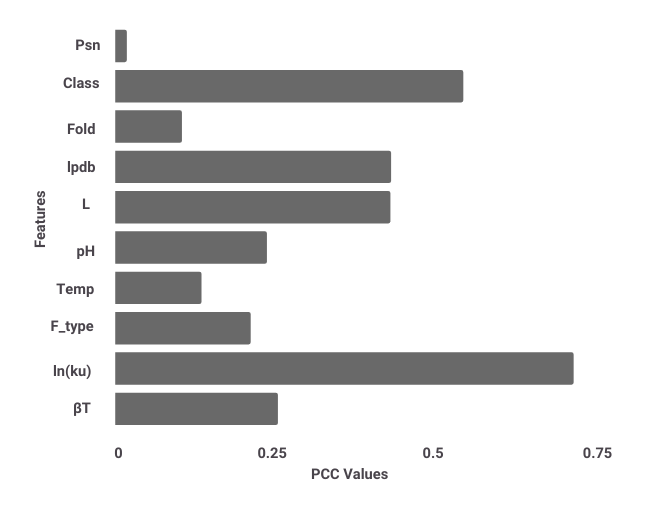}
  \caption{PFDB Feature Importance}
  \label{feature}
\end{figure}
As the number of features directly influences the efficiency of the model, finding the optimal number of features is essential to maximize the performance of the regressor while downsizing the number of feature samples and memory footprint. This would also reduce the computational cost for $\Delta$ and the overall pipeline.\\
Pearson's correlation coefficient \cite{sedgwick2012pearson} was used to determine the base relationship between the target variable ($ln(kf)$) and other feature sets. The correlation value (R) ranges from -1 to +1, where -1 (negative correlation) and +1 (positive correlation) signify strong correlation, while values closer to 0 signify weak or no correlation. R was computed, as shown in Equation ~\ref{eq2}, where $X_i$, $Y_i$ are individual datapoints and $X_m$, $Y_m$ is the mean of the X and Y dataset respectively, the absolute was taken to reside in the range of 0 to 1 for more uncomplicated depiction and understanding of the magnitude of correlation vectors as shown in Fig ~\ref{feature}. 


\begin{equation}
    R = \Bigg|\frac{\sum[(X_i - X_m)(Y_i - Y_m)]}{\surd[\sum(X_i - X_m)^2 * \sum(Y_i -Y_m)^2]}\Bigg|
    \label{eq2}
\end{equation}
\subsection{Training}
PFDB served as the primary dataset for training the regressor model (bonsai). PFDB contains a diverse collection of PFK information on both 2S and N2S proteins. A comprehensive train test strategy was implemented to assess the adaptability and generalizability of the trained model. The dataset was split into train and test as this approach ensured that the models were evaluated on unseen data, allowing for a robust assessment of their predictive capabilities. To optimize the model, MAE was iteratively reduced by tuning the hyperparameters. The process of testing the mini-batches and updating the parameters occurs concurrently by multi-threading. This approach reduced the time to compute results and significantly decreased MAE.  

The regressor was trained on different mini-batches, each comprising a unique feature subset. The training was conducted on $F_n = \{n = 2, 4, 5, 6, 7, 8, 9\}$, where $n$ represents the feature subset derived from Pearson's correlation coefficient analysis. Specifically, the batches $D^{\text{AA}}$, $D^{\text{BB}}$, $D^{\text{AB}}$, and $D^{\text{BA}}$ were trained on the prominent features $F_{\text{best}}^{\text{A}}$, $F_{\text{best}}^{\text{B}}$, $F_{\text{best}}^{\text{A}} \cup F_{\text{best}}^{\text{B}}$, and $F_{\text{best}}^{\text{A}} \cap F_{\text{best}}^{\text{B}}$, respectively. The features are chosen from $F_n$ as formulated in Equation ~\ref{eq1} and ~\ref{eq2}.
\begin{equation}\label{eq1}
   F_{n}^A = F^A \cup \{f : f_i \mid f_i \notin F^A \}
   \end{equation}
\begin{equation}\label{eq2}
    F_{n}^B = F^B \cup \{f : f_i \mid f_i \notin F^B \}
\end{equation}
\vspace{-0.5cm}
\begin{table}[ht]
\centering
\caption{Train Test stratergy}
\label{traintable}
\begin{tabular}{|l|l|l|}
\hline
\textbf{Testing} & $\boldsymbol{\operatorname{PFDB}(\mathrm{A})}$
 & $\boldsymbol{\operatorname{PFDB}(\mathrm{B})}$ \\
\hline
$\operatorname{PFDB}(\mathrm{A})$ & \begin{tabular}[c]{@{}l@{}}$D^{\text{AA}}$ \\ Training PFDB(A) \\ Testing PFBD(A)\end{tabular} & \begin{tabular}[c]{@{}l@{}}$D^{\text{BA}}$ \\ Training PFDB(B) \\ Testing PFBD(A)\end{tabular} \\
\hline
$\operatorname{PFDB}(\mathrm{B})$ & \begin{tabular}[c]{@{}l@{}}$D^{\text{AB}}$ \\ Training PFDB(A) \\ Testing PFBD(B)\end{tabular} & \begin{tabular}[c]{@{}l@{}}$D^{\text{BB}}$ \\ Training PFDB(B) \\ Testing PFBD(A)\end{tabular} \\
\hline
\end{tabular}
\end{table}
\subsection{Metrics}\label{metric}
The metrics actively used to evaluate the bonsai throughout this work are formulated below:

\begin{equation}
    \text{MAE} = \frac{1}{n} \sum_{i=1}^{n} |y_i - \hat{y}_i|
\end{equation}
\begin{equation}
    \text{MSE} = \frac{1}{n} \sum_{i=1}^{n} (y_i - \hat{y}_i)^2
\end{equation}
\begin{equation}
    R^2 = 1 - \frac{\sum_{i=1}^{n} (y_i - \hat{y}_i)^2}{\sum_{i=1}^{n} (y_i - \bar{y})^2}
\end{equation}
\\
$y_i$ = Actual kinetic value of the protein
\\
$\hat{y}_i$ = Predicted kinetic value of the protein
\\
$\bar{y}$ = Mean of the actual kinetic values, for $i = 1, 2, \ldots, n$.\\
Other evaluation metrics actively used in this work are inference time and model size. These metrics enable informed decisions regarding feature subset selection and model optimization.
\FloatBarrier
\section{Results and analysis}
\subsection{Bonsai results and analysis}
\begin{figure}
\centering
  \includegraphics[width=\columnwidth, height=5cm]{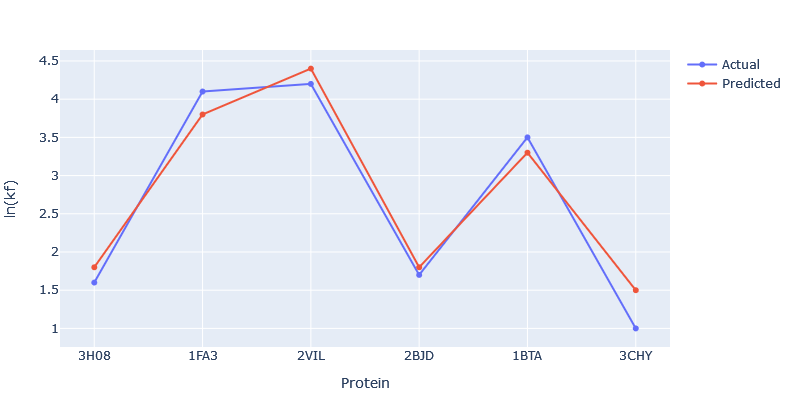}
  \caption{Actual vs predicted results for N2S protein kinetics}
  \label{feature}
\end{figure}
\begin{figure}
  \centering
  \includegraphics[width=\columnwidth, height=5cm]{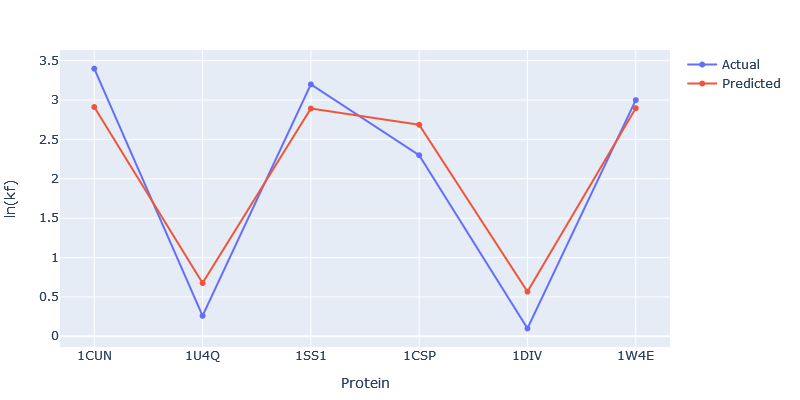}
  \caption{Actual vs predicted results for 2S protein kinetics}
  \label{feature}
\end{figure}
\begin{table*}
\centering
\caption{Results of models trained on all feature combinations}
\scalebox{1.15}{
\begin{tabular}{|c|c|c|c|c|c|c|c|}
\hline
\textbf{Data} & \textbf{Feature Subsets} & \textbf{MSE} & \textbf{MAE} & $\mathbf{R^2}$ & \textbf{Model Size (KB)} &\textbf{Bonsai IFT (ms)} &\textbf{Pipeline IFT (ms)} \\
\hline
$D_{AA}$ & $F_2^A$ & 1.4874 & 0.9573 & 0.889 & 0.486 & 5.2 &22.3 \\
\cline{2-8}
& $F_4^A$ & 0.9738 & 0.4518 & 0.927 & 0.608 & 5.2 &24.2 \\
\cline{2-8}
& $F_5^A$ & 1.0874 & 0.7283 & 0.912 & 0.528 & 5.2 &22.9 \\
\cline{2-8}
& $F_{6}^A$ & 1.0247 & 0.6598 & 0.919 & 0.531 & 5.2 &23.4 \\
\cline{2-8}
& $F_{8}^A$ & 1.1359 & 0.7653 & 0.903 & 0.496 &5.2 &22.5 \\
\cline{2-8}
& $F_{9}^A$ & 1.0475 & 0.6904 & 0.914 & 0.549 & 5.2 &23.7 \\
\hline
$D_{BB}$ & $F_2^B$ & 1.1076 & 0.7239 & 0.908 & 0.521 &5.2 & 22.6 \\
\cline{2-8}
& $F_4^B$ & 1.0421 & 0.7034 & 0.917 & 0.563 &5.2 & 24.1 \\
\cline{2-8}
& $F_5^B$ & 1.4579 & 0.9357 & 0.896 & 0.506 & 5.2 & 22.1 \\
\cline{2-8}
& $F_{6}^B$ & 1.1185 & 0.7159 & 0.904 & 0.542 &5.2 & 22.9 \\
\cline{2-8}
& $F_{8}^B$ & 1.0896 & 0.7693 & 0.911 & 0.562 & 5.2 & 23.9\\
\cline{2-8}
& $F_{9}^B$ & 0.8792 & 0.3179 & \textbf{0.934} & 0.627 & 5.2 & 24.7 \\
\hline
$D_{AB}$ & $F_{4}^A \cap F_{9}^B$ & 1.4298 & 0.9972 & 0.888 & 0.598 & 5.2 & 24.3 \\
\cline{2-8}
& $F_{4}^A \cup F_{9}^B$ & 1.1958 & 0.8198 & 0.901 & 0.633 &5.2 & 24.5 \\
\hline
$D_{BA}$ & $F_{4}^A \cap F_{9}^B$ & 1.1692 & 0.8196 & 0.916 & 0.615 &5.2 & 24.4 \\
\cline{2-8}
& $F_{4}^A \cup F_{9}^B$ & 1.2847 & 0.9271 & 0.895 & 0.592 &5.2 & 24.2 \\
\hline
\end{tabular}}
\end{table*}

As discussed above, an optimal feature subset heavily influences the performance of the bonsai model. The basis behind choosing an optimal feature subset is based on the performance, memory footprint, and inference time while availing the feature subsets. The detailed reasons for this are discussed in the upcoming sections.

\subsubsection{Best performance model}

The best scores for $D_{AA}$ were ascertained based on  MSE, MAE, and $R^2$. The scores were 0.9738,  0.4518, and 0.927 when $F_4^A$ was employed. Likewise, for $D_{BB}$, the best scores were attained by employing $F_9^B$, resulting in 0.8792, 0.3179, and 0.934. For $D_{AB}$, $F_{4}^A \cup F_{9}^B$ achieved best scores, 1.1958, 0.8198, and 0.901. Similarly, for $D_{BA}$, $F_{4}^A \cap F_{9}^B$ resulted in best scores, 1.1692, 0.8196, and  0.916. The better performance in $F_{4}^A \cap F_{9}^B$  compared to $F_{4}^A \cup F_{9}^B$ is due to the fact that $F_{4}^A \cap F_{9}^B$ uses fewer features than $F_{4}^A \cup F_{9}^B$. In a nutshell, if performance is of utmost importance, $F_{4}^A \cup F_{9}^B$ = $F_{9}^B$ represents the most favorable feature subset. 

\subsubsection{Best balance between  model size and performance}
The usual trend of better performance was linear with the number of features until it reached a point. This goes the same for model size and inference time (IFT) as well. $F_2^A$ feature subset is the most compact and computes in less time comparatively, but, when taking performance into account, other feature subsets deliver better performance. If a feature subset that strikes a balance between model size and performance had to be selected, $F_6^A$ or $F_6^B$ would be the best pick. The resulting increase in model size when compared to $F_2^A$ is due to the increase in the number of features in $F_6^B$. Furthermore, bonsai was 4.8\% more accurate, 327$\times$ more compact in terms of model size, and 7.3$\times$ faster when compared to prior SoA implementation \cite{NITHIYANANDAM2023106436}.
\subsection{Standard ML models}
Several regressor models, such as Decision Tree (DT), Random Forest (RF), XGBoost (XGB), and LightGBM (LGBM), were trained in the same environment as the bonsai model to compare and contrast the difference in performance among them. Since bonsai employs tree-based regression, five tree-based regressors were chosen for comparison. These models were trained on the most prominent feature set $F_{9}^B$, as bonsai is designed to minimize computational resources, taking the best balance model into account would not be an appropriate comparison. Bonsai's ability to achieve such high-performance gains while maintaining its compactness and accuracy sets it apart. The best regressor model (LGBM), in terms of performance, was evidently outperformed by the bonsai model, which consumed only 9\% of the memory consumed by LGBM while being 6\% more accurate when evaluated using R2 score. Bonsai was also 156x faster than DT, the model with the lowest inference time (8.736 sec).  

\subsection{Pipeline Analysis}\FloatBarrier
Even though the bonsai model's performance and memory footprint is carefully analyzed and reported, analysis of the pipeline is also fairly significant. As depicted in Fig ~\ref{eq2}, $\Delta$ consumes the most time (71\%), out of which normalization and standardization ($\beta$) take up to 37\% and feature extractor consumes 34\% individually. The regressor consumes 21\% of the pipeline inference time, while H takes up 4\%-9\% of inference time, which is comparatively insignificant. This pipeline was constructed primarily to consume minimal computational resources while being highly accurate and fast (24.7 ms). This makes the pipeline highly compatible for future integration onto fold-specific prediction frameworks and systems. 

\renewcommand{\arraystretch}{0.8}
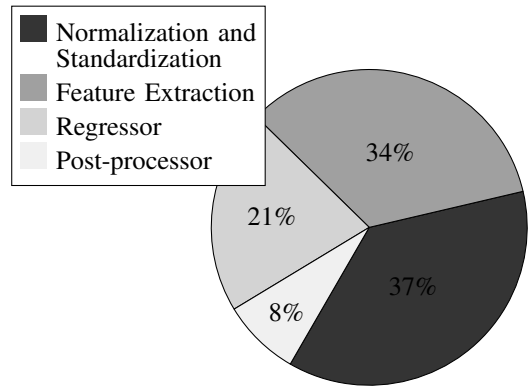
\begin{figure}
    \centering
   \begin{tikzpicture}[scale=0.7]

  \definecolor{color1}{RGB}{52,52,52}      
\definecolor{color2}{RGB}{160,160,160}  
\definecolor{color3}{RGB}{211,211,211}  
\definecolor{color4}{RGB}{240,240,240}   

  \pie[
    /tikz/every pin/.style={align=center},
    every only number node/.style={text=white},
    text=,
    rotate=240,
    explode=0,
    color={color1,color2,color3,color4},
    ,
    style=,
    before number=
  ]{
    37/,
    34/,
    21/,
    8/
}

  \node[rectangle,draw,fill=white,anchor=west,minimum width=2cm, minimum height=1.5cm] at (-6.8,2.5) {\begin{tabular}{@{}r@{ }l@{}}
    \textcolor{color1}{\rule{10pt}{10pt}} & Normalization and \\ & Standardization \\
    \textcolor{color2}{\rule{10pt}{10pt}} & Feature Extraction \\
    \textcolor{color3}{\rule{10pt}{10pt}} & Regressor \\
    \textcolor{color4}{\rule{10pt}{10pt}} & Post-processor \\
  \end{tabular}};
\end{tikzpicture}
\captionsetup{justification=centering}
\caption{Usage of computational resources}
\end{figure}
The above chart was plotted for the best performing model which was trained on $F_{9}^B$. Between the models trained on $F_{9}^B$ and $F_{8}^A$, $F_{9}^B$ was chosen so as to understand and illustrate the computational needs of the best regressor. $F_{8}^A$ was not considered as it does not exhibit the best performance but only the best balance between the considered factors, model size, performance, and inference time.
\FloatBarrier
\section{Conclusion}
This research presented a novel optimized pipeline for accurate protein folding kinetics prediction using protein-specific parameters as its input. The proposed pipeline significantly outperformed the standard ML models in terms of accuracy, memory footprint, and inference time. The optimized pipeline also showcases great adaptability and predictive capabilities when compared with SoA frameworks. The implemented model achieved an MSE of 0.8792, MAE of 0.3179, and a $R^{2}$ score of 0.934 in an inference time of 5.2ms while consuming 396x lesser memory than other SoA systems. The implemented pipeline can be integrated into future protein folding prediction systems for accurate and fast PFK prediction, as it facilitates PFK prediction while reducing computational resources and latency.
\bibliographystyle{ieeetr}
\bibliography{reference}
\end{document}